\documentclass[10pt,twocolumn,letterpaper]{article}

\usepackage{cvpr}              
\usepackage{algorithm}
\usepackage{algorithmic}
\usepackage{amssymb}
\usepackage{amsmath}
\usepackage{svg}
\usepackage{booktabs}
\usepackage{multirow}
\usepackage{xcolor}
\usepackage{footmisc}

\definecolor{cvprblue}{rgb}{0.21,0.49,0.74}
\usepackage[pagebackref,breaklinks,colorlinks,allcolors=cvprblue]{hyperref}

\def\paperID{8903} 
\def\confName{CVPR}
\def\confYear{2026}

\title{HDW-SR: High-Frequency Guided Diffusion Model based on Wavelet Decomposition for Image Super-Resolution}

\author{
Chao Yang\footnotemark[1],  
Boqian Zhang\footnotemark[1], 
Jinghao Xu,
Guang Jiang \footnotemark[2]\\
Xidian University\\
{\tt\small 
  \{24011211295, 24011211294, 23011210789\}@stu.xidian.edu.cn, 
}\\
{\tt\small gjiang@mail.xidian.edu.cn}
}

\begin{document}
\maketitle
\begin{abstract}



Diffusion-based methods have shown great promise in single image super-resolution (SISR); however, existing approaches often produce blurred fine details due to insufficient guidance in the high-frequency domain. To address this issue, we propose a High-Frequency Guided Diffusion Network based on Wavelet Decomposition (HDW-SR), which replaces the conventional U-Net backbone in diffusion frameworks. Specifically, we perform diffusion only on the residual map, allowing the network to focus more effectively on high-frequency information restoration. We then introduce wavelet-based downsampling in place of standard CNN downsampling to achieve multi-scale frequency decomposition, enabling sparse cross-attention between the high-frequency subbands of the pre-super-resolved image and the low-frequency subbands of the diffused image for explicit high-frequency guidance. Moreover, a Dynamic Thresholding Block (DTB) is designed to refine high-frequency selection during the sparse attention process. During upsampling, the invertibility of the wavelet transform ensures low-loss feature reconstruction. Experiments on both synthetic and real-world datasets demonstrate that HDW-SR achieves competitive super-resolution performance, excelling particularly in recovering fine-grained image details. The code will be available after acceptance.
\end{abstract}
\footnotetext[1]{These authors contributed equally to this work.}    
\footnotetext[2]{Corresponding authors.} 
\section{Introduction}


Single Image Super-Resolution (SISR) aims to reconstruct a high-resolution (HR) image from a low-resolution (LR) input. This task has been widely applied in low-level vision. However, due to the resolution limitations of the LR image, certain details may be blurred or lost, making accurate restoration of fine details a critical challenge for SISR.

\begin{figure}[t]
\centering
\includegraphics[width=1.0\columnwidth]{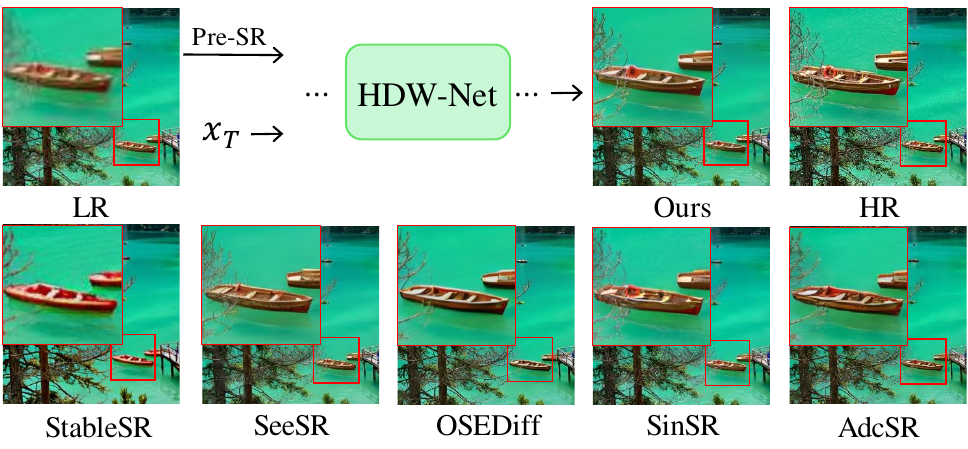} 
\caption{Existing methods achieve good performance in restoring the overall structure, but still have deficiencies in detail reconstruction (e.g. the lifebuoy on the boat).}
\label{fig1}
\end{figure}
In recent years, Generative Models have achieved advances in the SISR task, particularly in restoring high-frequency details and producing high-quality images. Compared to traditional regression methods, GAN-based methods in SISR\cite{park2023perception} can explicitly constrain the generator’s output to match the natural image distribution, thereby synthesizing more realistic SR images. Moreover, GANs readily incorporate perceptual and feature-matching losses to further enhance subjective texture fidelity. Nevertheless, training GAN-based approaches is difficult, and the convolutional upsampling operations often introduce conspicuous checkerboard artifacts in the generated HR images \cite{yue2023resshift}.  


Compared to GANs, diffusion models can achieve better recovery of high-frequency details and texture information during the image generation process through a step-by-step denoising approach. Currently, some diffusion-based methods \cite{li2022srdiff,niu2023cdpmsr,yue2023resshift,wang2024sinsr} typically generate HR images from random noise through multiple denoising steps. This approach not only prolongs convergence, but also limits the model's ability to focus on fine-grained details, leading to texture details loss. Subsequent methods, such as Pre-trained Diffusion Models \cite{niu2024acdmsr,moser2025dynamic,wang2024exploiting} and Residual Learning Diffusion Models \cite{yue2023resshift,shi2025multi}, have shown improvements in detail recovery. However, these approaches still exhibit deficiencies in recovering local details and handling multi-scale features due to their reliance on global image features and the absence of high-frequency prior guidance.


To overcome these limitations, we propose a High-Frequency Guided Diffusion Model based on Wavelet Decomposition for Image Super-Resolution(HDW-SR). Since wavelet decomposition can losslessly separate an image into low- and high-frequency components, it avoids the high-frequency detail loss commonly caused by traditional CNN downsampling. Therefore, the model integrates wavelet transforms to extract multi-scale features and uses a pre-super-resolved (PreSR) image, which is resolved from LR image, to provide high-frequency guidance during diffusion. Specially, we first use PreSR image to generate a residual map based on the difference between PreSR image and the HR image. Subsequently, diffusion is performed based on the residual map. In the diffusion process, we replace the traditional U-Net with the High-Frequency Guided Diffusion by Wavelet Decomposition Network (HDW-Net), which utilizes the high-frequency prior guidance provided by PreSR image during the diffusion.


HDW-Net consists of two components: the High-Frequency Extraction Network (HE-Net) and the High-Frequency Augmentation Network (HA-Net). HE-Net serves as a feature extraction network that performs multi-level wavelet decomposition on the PreSR image to extract high-frequency components at different scales. By employing a U-Net–like architecture optimized with reconstruction loss, HE-Net ensures that the extracted wavelet components are precise and robust, providing reliable high-frequency guidance for the subsequent HA-Net. In HA-Net, the denoised image is processed through wavelet decomposition, which effectively separates low- and high-frequency components while preserving spatial structures, thus enhancing the model’s detail perception and reconstruction accuracy\cite{liu2018multi}. The low-frequency components obtained from wavelet decomposition and the high-frequency components extracted by HE-Net are fed into an encoder based on Dynamic Focused Attention (DFA), where sparse cross-attention computation guides the high-frequency prior. The combination of residuals and high-frequency components effectively strengthens the network's ability to restore fine details, avoiding excessive noise or distortion.


To precisely select essential elements from the similarity matrix and enhance high-frequency guidance quality, we propose the Dynamic Thresholding Block (DTB), an enhancement over the conventional Top-K mechanism. Observing that elements in the normalized similarity matrix of the cross-attention often exhibit a bimodal distribution, DTB dynamically computes thresholds based on intra-class and inter-class variances, allowing more precise element selection. This method significantly refines the network's focus on high-frequency details and accelerates computational efficiency.

The main contributions of our work are as follows:




\begin{itemize}
\item We propose HDW-Net, a network structure that integrates wavelet transform and DFA-based high-frequency feature encoding within the diffusion framework. By replacing traditional CNN-based sampling with wavelet transforms and using the high-frequency information from the PreSR image as guidance for the diffusion process, our approach significantly enhances detail reconstruction accuracy.
\item Within DFA, we propose the DTB, a module that dynamically determines data selection thresholds by analyzing the variance in similarity matrices, enabling more precise and adaptive selection of key elements.
\item Extensive experiments have demonstrated that our model achieves superior performance and metrics, particularly excelling in detail restoration.
\end{itemize}
\section{Related Work}



Deep learning-based methods have become the mainstream approach for SISR. The application of CNNs and Transformer \cite{vaswani2017attention} has significantly enhanced SISR performance \cite{lim2017enhanced, zhang2021two, li2023efficient, yoo2023enriched, zhang2024hit}. The Residual Dense Network (RDN) \cite{zhang2018residual}, by incorporating local and global feature fusion mechanisms within multiple residual dense blocks, effectively exploits multi-level information. The Progressive Focused Transformer (PFT) \cite{long2025progressive},  utilizing a sparse attention mechanism, improves the SISR performance while reducing computational complexity. Furthermore, significant advancements have been made in generative-model-based SISR methods, which can be broadly categorized into GAN-based approaches \cite{zhang2019ranksrgan, liang2022details, park2023perception} and diffusion-based methods \cite{li2022srdiff, wang2024enhancing, chao2025lfsrdiff}.



\subsection{Diffusion-based SISR}


Recently, diffusion models have made significant progress in the field of image generation \cite{dhariwal2021diffusion, huang2024wavedm} and have gradually been applied to SISR reconstruction tasks, Gendy's work \cite{gendy2025diffusion} provides a summary of the application of diffusion models in image super-resolution. Methods based on generative priors have achieved notable success with Stable Diffusion \cite{rombach2022high, podell2023sdxl}. For instance, ResShift \cite{yue2023resshift} effectively reduces denoising steps by shortening the Markov chain, while ResDiff \cite{shang2024resdiff} significantly improves the reconstruction accuracy of high-frequency details within a frequency-domain-guided framework. Single-step diffusion-based SISR  methods \cite{wu2024one, yang2024pixel, chen2025adversarial} achieve performance comparable to multi-step methods with far fewer sampling steps. However, existing methods often still suffer from issues of excessive smoothing or artifact introduction in high-frequency regions, necessitating more refined attention mechanisms and prior fusion strategies to enhance SISR performance.

\subsection{Wavelet transform for SISR}

Wavelet decomposition \cite{daubechies1988orthonormal} combined with deep learning techniques \cite{xue2020wavelet, hsu2022detail, aloisi2024wavelet, zhang2022perception} has made significant progress in the field of SISR. The Lifting Wavelet-Guided Diffusion Model \cite{parida2025lifting} improves image super-resolution accuracy and efficiency by restoring details in the wavelet domain. However, its performance is limited by the diffusion model's struggle to distinguish between details and noise during denoising, affecting reconstruction quality. DiWa \cite{moser2024waving} transfers the diffusion process to the wavelet domain. But frequency-domain diffusion depends on input image quality, and the lack of residual learning hinders precise detail recovery. Therefore, our work applies the wavelet transform to the residual diffusion process,  utilizing the transform to inject multi‑scale high‑frequency information, enabling more accurate recovery of high-frequency details and thus achieving SISR in the image domain.

\section{Method}

This chapter elaborates on the HDW-SR architecture. In Section~\ref{Section 1}, we present the overall structure of the HDW-SR. Section~\ref{Section 2} discusses the design of the HDW-Net. Finally, in Section~\ref{Section 3}, we provide a detailed explanation of the encoder design based on the DFA mechanism with dynamic thresholding.
 
\subsection{Prior-based Residual Diffusion Network}
\label{Section 1}

\begin{figure}[t]
\centering
\includegraphics[width=1.0\columnwidth]{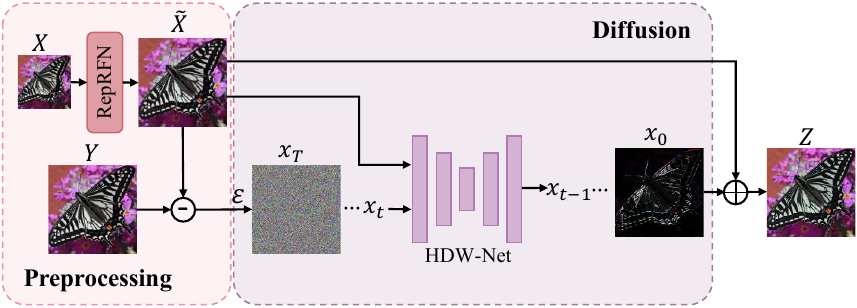} 
\caption{Overview of Prior-based Residual Diffusion Network.}
\label{fig1}
\end{figure}



The detailed design of the network is shown in Figure~\ref{fig1}. We consider image pairs $(X_i, Y_i) \in \mathcal{D}$, where $X_i \in \mathbb{R}^{H \times W \times 3}$ is the LR image and $Y_i \in \mathbb{R}^{H \times W \times 3}$ is the HR image.To make the network more focused on learning high-frequency information and reduce training costs, we perform diffusion on the residuals: First, the lightweight CNN network RepRFN \cite{deng2023reparameterized} is used to preprocess $X_i$, and the PreSR image $\tilde{X}_i \in \mathbb{R}^{H \times W \times 3}$ is subtracted from $Y_i$ to obtain the residual image $\Delta X_i \in \mathbb{R}^{H \times W \times 3}$ which narrows the dynamic range of the signal that the model must learn. By progressively adding noise to $\Delta X_i$, we obtain the pure noise sample $x_T$. During the subsequent reverse denoising process, the PreSR $\tilde{X}_i$ serves as a guidance signal and, together with the noise sequence $x_t, \, t=0 \dots T$, is fed into HDW-Net to predict the residual map $\Delta X_{\theta, i}$. Finally, adding $\Delta X_{\theta, i}$ to $\tilde{X}_i$ yields the super-resolved image $Z_i \in \mathbb{R}^{H \times W \times 3}$.

\subsection{HDW-Net}
\label{Section 2}


HDW-Net replaces CNN-based downsampling and upsampling operations with wavelet transforms to process the noise $x_t$ and the PreSR $\tilde{X}_i$. The high-frequency wavelet components decomposed from $\tilde{X}_i$ are employed as guidance throughout the diffusion process. As shown in Figure ~\ref{fig: diffusion process}, HDW-Net consists of two sub-networks: the HE-Net and the HA-Net.



\subsubsection{HE-Net} 
\begin{figure*}[t]
\centering
\includegraphics[width=0.9\textwidth]{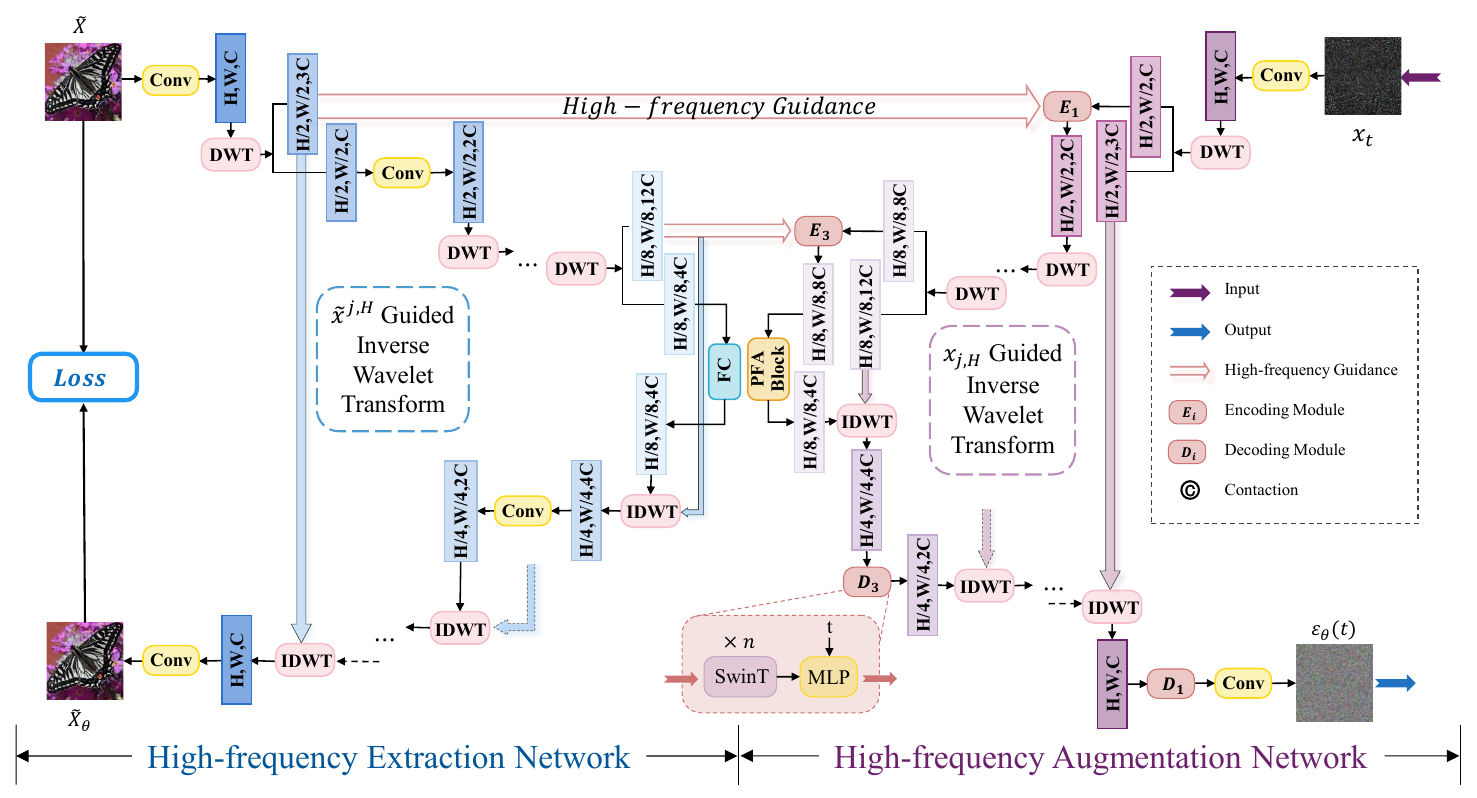} 
\caption{Overview of HDW-Net. HE-Net extracts the high-frequency component of $\tilde{X}$ and feeds it into the HA-Net on the right. The HA-Net encoder then computes cross-attention between high- and low-frequency components to extract detailed features, completing the diffusion process.}
\label{fig: diffusion process}
\end{figure*}

As shown in the left panel of Figure~\ref{fig: diffusion process}, the input image $\tilde{X}$ undergoes multi-level downsampling using the Haar wavelet \cite{haar1909theorie}. In the $j$-th sampling step, the image $\tilde{x}^{j-1}, \; j=1,2,3$ is decomposed into four subbands based on the Haar wavelet basis.

\begin{equation}
\begin{split}
\text{2D-DWT}\bigl(\tilde{x}^{j-1}\bigr)=\tilde{x}^{j,LL}, \tilde{x}^{j,LH}, \tilde{x}^{j,HL}, \tilde{x}^{j,HH},
\end{split}
\end{equation}


The term $\tilde{x}^{j,LL}$ represents the low-frequency component that captures the structural features of the image. After a convolutional operation adjusts its channel dimension, the resulting $\tilde{x}^j$ is used for next-level wavelet downsampling. The set ${\tilde{x}^{j,LH},\,\tilde{x}^{j,HL},\,\tilde{x}^{j,HH}}$ corresponds to the high-frequency components, collectively denoted as $\tilde{x}^{j,H}$ in the following discussion. These components extract fine-grained details of the image and serve as high-frequency guidance, fed into the corresponding layers of HA-Net during the denoising process. Repeating this procedure iteratively, the final low-frequency component $\tilde{x}^{3,LL}$ is passed through a fully connected (FC) layer to produce the enhanced feature $\tilde{x}_{\theta}^{3,LL}$, which initializes the low-frequency wavelet component for the upsampling process. Subsequently, the low-frequency component at each level, together with its corresponding high-frequency component, is progressively reconstructed through the inverse wavelet transform: $\tilde{x}_{ \theta}^{j-1} = \text{2D-IDWT}(\tilde{x}_{\theta }^{j,LL}, \tilde{x}^{j,H})$.


After completing the final upsampling step, a CNN is used to obtain the output $\tilde{X}_{\theta}$. The loss function $\mathcal{L}_{HE}$ is applied to constrain the HE-Net, ensuring high-quality restoration of $\tilde{X_i}$ and providing accurate high-frequency guidance for the HA-Net. The loss $\mathcal{L}_{HE}$ is defined as follows:
\begin{align}
\mathcal{L}_{HE} = \left\|\tilde{X} - \tilde{X}_{\theta}\right\|_2 + \left\|\tilde{X} - \tilde{X}_{\theta}\right\|_1,
\end{align}

\subsubsection{HA-Net}


 The diffusion process is illustrated on the right side of Figure~\ref{fig: diffusion process}, where $x_t$ is fed into HA-Net and passed through convolution layers to align its channel dimensions with those of HE-Net, yielding $x_t^0$. Similarly, for the $j$-th layer sampling, Haar wavelet downsampling is first applied:

\begin{equation}
\begin{split}
\text{2D-DWT}\bigl(x^{j-1}_t\bigr)= x^{j,LL}_t, x^{j,LH}_t, x^{j,HL}_t , x^{j,HH}_t.
\end{split}
\end{equation}


From $x^{j-1}_t \, (j=1,2,3)$ , four subbands are obtained through wavelet downsampling, including the low-frequency subband $x^{j,LL}_t$ and the high-frequency subband $x^{j,H}_t$. The downsampling mechanism of the wavelet transform prevents the high-frequency information loss commonly caused by traditional pooling, particularly in regions rich in edges or textures. Compared with other frequency-domain methods, the wavelet transform provides a joint space–frequency localized representation, which is more suitable for optimizing high-frequency details in super-resolution tasks.
The low-frequency subband $x^{j,LL}_t$ and the corresponding high-frequency component $\tilde{x}^{j,H}$ from HE-Net are jointly fed into the encoding module $E_j$ (to be detailed in Section~\ref{Section 3}). The PreSR $\tilde{X}_i$ provides high-frequency guidance for the diffusion process, formulated as $x^{j+1}_t = \mathcal{E}(x^{j,LL}_t, \; \tilde{x}^{j,H}) $ . Repeating this process iteratively, the output $x^3_t$ of the final encoding module is passed through the Progressive Sparse Attention (PFA) module to obtain $x^{3,LL}_{t,\theta}$.


In the upsampling stage, we use $x_t^{j,H}$ instead of $\tilde{x}^{j,H}$ as the high-frequency component for the inverse wavelet transform. This design choice stems from the observation that, at the early diffusion stages, the high-frequency prior of $\tilde{X}$ provides strong guidance for detail restoration in $x_t$; however, as $t$ gradually decreases toward 0, $x_t$ becomes increasingly rich in high-frequency details. Excessive reliance on $\tilde{X}$ at this point would amplify pseudo-details and distort textures, thereby hindering further refinement of high-frequency quality.
The upsampling process is then repeated as follows: $x_{t, \theta}^{j-1}=\text{2D-IDWT}\bigl(x_{t, \theta}^{j,LL}, x_t^{j,H})$ , $x_{t, \theta}^{j-1,LL} = \mathcal{D}(x_{t, \theta}^{j-1})$, where $\mathcal{D}$ denotes the decoder composed of $n$ Swin-T layers and an MLP block (the former enhances high-dimensional features, while the latter embeds the timestep $t$). The final predicted diffusion noise $\varepsilon_\theta(x_t, t)$ is then used to compute the HA-Net loss $\mathcal{L}_{HA}$:

\begin{align}
    \mathcal{L}_{HA}=\mathrm{E}_\theta\bigl\|\,
            \varepsilon_t -\varepsilon_\theta (x_t,\;t)
        \bigr\|^2.
\end{align}


We define the overall loss function of the network as $\mathcal{L}$:

\begin{align}
    \mathcal{L}=\beta \mathcal{L}_{HE}+(1-\beta)\mathcal{L}_{HA},
\end{align}
Here, $\beta$ is a hyperparameter, and in this network, we set $\beta = 0.2$.

\begin{figure*}[t]
\centering
\includegraphics[width=0.9\textwidth]{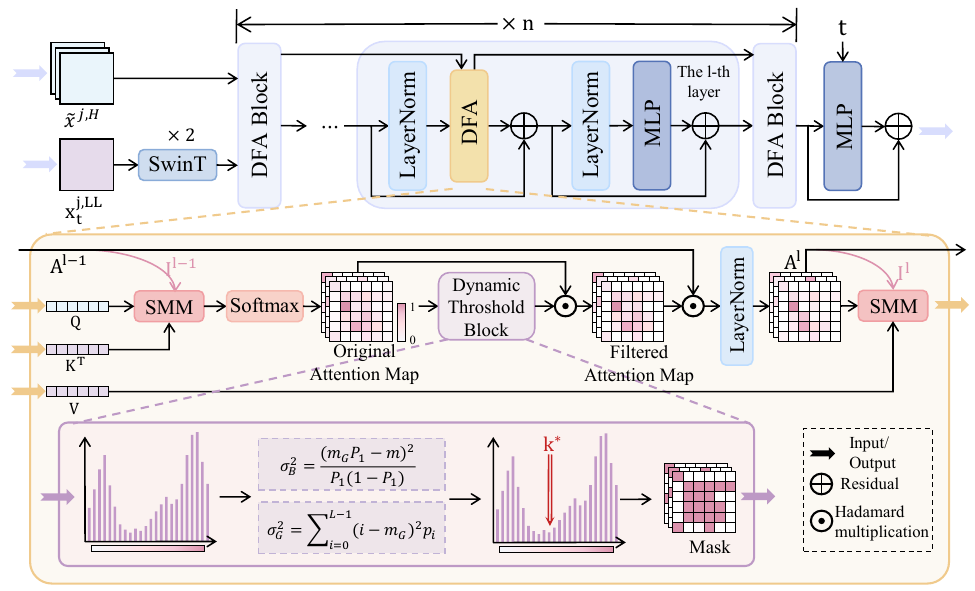} 
\caption{Overview of the DFA‑based encoder: DFA performs sparse cross‑attention between low‑ and high‑frequency wavelet components, while DTB dynamically selects K via inter‑class and intra‑class variances, supplanting Top‑K.}
\label{fig DFA}
\end{figure*}

\subsection{DFA-Based Encoder}
\label{Section 3}


Motivated by PFA\cite{long2025progressive}, we design a DFA-Based encoder using a dynamic‑threshold sparse attention mechanism for efficient high‑frequency guidance. In $E_j$, the $x_t^{j,LL}$ is first enhanced through two layers of Swin-Transformer \cite{liu2021swin}. Subsequently, the enhanced low-frequency feature (used as $Q$) and the corresponding high-frequency guidance feature $\tilde{x}^{j,H}$ (used as both $K$ and $V$) are jointly fed into the DFA module to perform cross-attention between high- and low-frequency representations.


Here, we introduce a sparse matrix multiplication (SMM), denoted as "$\Psi$" to reduce computational complexity; it also performs sparse indexing based on the previous layer’s attention indices $I^{l-1}\in\mathbb{R}^{N\times N}$ :
\begin{align}
A_{\mathrm{oam}}^{l} = \operatorname{Softmax}\bigl(\Psi(Q^{l},(K^{l})^{\top},I^{l-1})\bigr)
\end{align}

The update rule for the sparse indexing matrix is $I^{l} = \operatorname{Sign}(A^{l})$.


Due to the significant difference between $x^{j,LL}_t$ and $\tilde{x}^{j,H}$, the values in $A_{\mathrm{oam}}^{l}$ often exhibit a bimodal distribution. Based on this observation, we propose the Dynamic Thresholding Block (DTB), which adaptively determines the Top-$K$ threshold in sparse attention. Specifically, the element values $a_{i,j}$ in $A_{\mathrm{oam}}^{l}$ are organized into a histogram over the range $[0,1]$ with intervals of $1/512$, as shown in Figure~\ref{fig DFA}.


We divide $a_{i,j}$ into two categories based on a variable threshold $T(k)=k$: class $C_1$ contains the elements within the interval $[0,k]$, while the remaining elements are assigned to class $C_2$. Following the computation of intra-class variance $\sigma_c^2(k)$ and inter-class variance $\sigma_B^2(k)$ in image thresholding\cite{4310076}, the optimal threshold $k^*$ is determined as $k^* = \arg\max_{k} \,\sigma_{B}^{2}(k)$ .




Based on the optimal threshold $k^*$, the elements in $A^{l}_{\mathrm{oam}}$ greater than $k^*$ are set to 1, and the remaining elements are set to 0, resulting in the dynamic MASK. By element-wise multiplication with MASK, the elements in $A^{l}_{\mathrm{oam}}$ are selectively filtered in a more flexible manner, yielding $A^{l}_{\mathrm{fam}}$.

The attention map for the current layer is:
\begin{align}
A^{l} = \operatorname{Norm}(A^{l-1} \odot A^{l}_{\mathrm{fam}}).
\end{align}
By incorporating the previous layer’s attention map $A^{l-1}$ from the previous layer, high‑frequency guiding information is propagated through successive multiplications. Subsequently, the output of the current layer is computed using $\Psi$: 
\begin{align}
O^l=\Psi(A^l,V^l,I^l).
\end{align}


This completes a sign DFA operation. By iterating the DFA block n times, cross-attention guidance is established from  $\tilde{x}_i^{j,H}$ to $x^{j,LL}_t$, thereby fulfilling the functionality of the encoder module. Thanks to dynamic-threshold sparse attention and layer-wise weights propagation, the DFA-based encoder reduces computational complexity while accurately preserving and enhancing cross-layer consistent high-frequency key information. This facilitates the extraction of more discriminative fine-grained features for subsequent reconstruction.

\begin{table*}[ht]
\centering
\scriptsize
\setlength{\tabcolsep}{8pt}
\begin{tabular}{lccccccccccc}
\toprule
Datasets & Methods & Publication & PSNR$\uparrow$ & SSIM$\uparrow$ & LPIPS$\downarrow$ & DISTS$\downarrow$ & NIQE$\downarrow$ & CLIPIQA$\uparrow$ & MUSIQ$\uparrow$ \\
\midrule
\multirow{8}{*}{DIV2K} 
 & ResShift & NeurIPS'2023 & \textbf{24.69} & \textbf{0.6175} & 0.3374 & 0.2215 & 6.82 & 0.6089 & 60.92\\
 & StableSR & IJCV'2024 & 23.31  & 0.5728  & 0.3129 & 0.2138 & 4.76 & 0.6682 & 65.63 \\
 & SeeSR & CVPR'2024 & 23.71 & 0.6045 & 0.3207 & 0.1967 & 5.83 & \underline{0.6857} & 68.49 \\
 & SinSR & CVPR'2024 & 24.43 & 0.6012 & 0.3262 & 0.2066 & 6.02 & 0.6499 & 62.80 \\
 & OSEDiff & NeurIPS'2024 & 23.72 & 0.6108 & 0.2941 & 0.1976 & 4.71 & 0.6693 & 67.97 \\
 & AdcSR  & CVPR'2025 & 23.62 & 0.6052 & 0.3062 & 0.1994 & 4.82 & 0.6594 & 69.37\\
 & DiT-SR & AAAI'2025& 24.31 & 0.6074 & \underline{0.2913} & \underline{0.1956} & \underline{4.55} & 0.6711 & \underline{69.47}\\
 & Ours   & - & \underline{24.52} & \underline{0.6162} & \textbf{0.2823} & \textbf{0.1934} & \textbf{4.43} & \textbf{0.6937} & \textbf{69.68} \\
\midrule
\multirow{8}{*}{RealSR} 
 & ResShift  & NeurIPS'2023 & \textbf{26.31} & \underline{0.7411} & 0.3489 & 0.2498& 7.27 & 0.5450 &58.10 \\
 & StableSR & IJCV'2024 & 24.69 & 0.7052 & 0.3091 & 0.2167 & 5.76 & 0.6195 & 65.42 \\
 & SeeSR  & CVPR'2024 & 25.33 & 0.7273 & 0.2985 & 0.2213 & 5.38 & 0.6204 & \underline{69.37} \\
 & SinSR & CVPR'2024 & \underline{26.30} & 0.7354 & 0.3212 & 0.2346 & 6.31 & 0.6594 & 60.41 \\
 & OSEDiff & NeurIPS'2024 & 25.15 & 0.7341 & 0.2921 & \underline{0.2128} & \underline{5.37} & 0.6693 & 69.09 \\
 & AdcSR  & CVPR'2025 & 24.92 & 0.7284 & 0.3006 & 0.2216 & 5.61 & 0.6594 & 69.17\\
 & DiT-SR & AAAI'2025 & 25.31 & 0.7337 & \underline{0.2863} & 0.2181 & \textbf{5.36} & \textbf{0.6961} & 65.83\\
 & Ours & - & 25.71 & \textbf{0.7428} & \textbf{0.2672} & \textbf{0.2044} & 5.39 & \underline{0.6702} & \textbf{70.10} \\
\midrule
\multirow{8}{*}{DrealSR} 
 & ResShift & NeurIPS'2023 & \textbf{28.45} & 0.7632 & 0.3489 & 0.2498 & 8.28 & 0.5259 & 49.86 \\
 & StableSR & IJCV'2024 & 28.04 & 0.7460 & 0.3354 & 0.2287 & 6.51 & 0.6167 & 58.50 \\
 & SeeSR & CVPR'2024 & 28.26 & 0.7698 & 0.3197 & 0.2306 & 6.52 & 0.6672 & 64.84 \\
 & SinSR & CVPR'2024 & 28.41 & 0.7495 & 0.3741 & 0.2488 & 7.02 & 0.6367 & 55.34 \\
 & OSEDiff & NeurIPS'2024 & 27.92 & \textbf{0.7835} & \underline{0.2968} & 0.2165 & 6.49 & \underline{0.6963} & 64.65 \\
 & AdcSR  & CVPR'2025 & 28.10 & 0.7726 & 0.3046 & 0.2200 & 6.45 & 0.6849 & \underline{66.26}\\
 & DiT-SR & AAAI'2025 & 28.17 & 0.7792 & 0.3015 & \underline{0.2164} & \underline{6.31} & 0.6920 & 65.74\\
 & Ours & - & \underline{28.43} & \underline{0.7824} & \textbf{0.2960} & \textbf{0.2153} & \textbf{6.20} & \textbf{0.6970} & \textbf{66.36} \\
\bottomrule
\end{tabular}
\caption{Quantitative comparison among the state-of-the-art DM-based SR methods on synthetic and real-world test datasets. The best and second-best results are \textbf{in bold} and \underline{underlined}, respectively.}
\label{tab: quantitative comparison}
\end{table*}

\section{Experiments}
\subsection{Experiments Settings}
\subsubsection{Experiments Environment.}

The proposed model was trained on a workstation equipped with an Intel Core i9-14900 CPU and dual NVIDIA RTX 4090 GPUs, running Ubuntu 22.04. The implementation utilized PyTorch 2.0.1 with CUDA 11.8. For training, low-resolution (LR) and high-resolution (HR) image pairs from the DIV2K and LSDIR datasets were employed, incorporating a 4$\times$ upsampling factor. In the network architecture, the DFA module was repeated [2, 4, 4] times across its three respective stages, while the Swin-T decoder consisted of [4, 6, 6] repeated layers. Model optimization was conducted using the Adam optimizer with an initial learning rate of $1 \times 10^{-4}$, over a total of 100,000 training iterations.

\subsubsection{Compared methods.}

We compare HDW-SR with multi-step DM-based methods StableSR \cite{wang2024exploiting} , ResShift \cite{yue2023resshift}, SeeSR \cite{wu2024seesr}, one-step DM-based methods SinSR \cite{wang2024sinsr}, OSEDiff \cite{wu2024one}, AdcSR \cite{chen2025adversarial} and the DiT‑based method DiT-SR \cite{cheng2025effective}. All comparative results are obtained using officially released codes or models.
\subsubsection{Test datasets and evaluation metrics.}

Following previous work\cite{wu2024one,wu2024seesr}, we evaluate all methods on both synthetic and real-world datasets. The synthetic dataset consists of 3,000 images of size 512×512 cropped from DIV2K\cite{agustsson2017ntire}. The real‑world images are center‑cropped from RealSR\cite{cai2019toward} and DrealSR\cite{wei2020component}. PSNR and SSIM\cite{wang2004image}are used to measure the fidelity of SR images; LPIPS\cite{zhang2018unreasonable} and DISTS\cite{ding2020image} are used to assess the perceptual quality of SR; NIQE\cite{mittal2012making}, CLIPIQA\cite{wang2023exploring}, and MUSIQ\cite{ke2021musiq} evaluate image quality without reference images. 

\subsection{Comparisons with State-of-the-Arts}

\subsubsection{Comparison Results on the Synthetic DIV2K Dataset}
\begin{figure*}[t]
\centering
\includegraphics[width=0.9\textwidth]{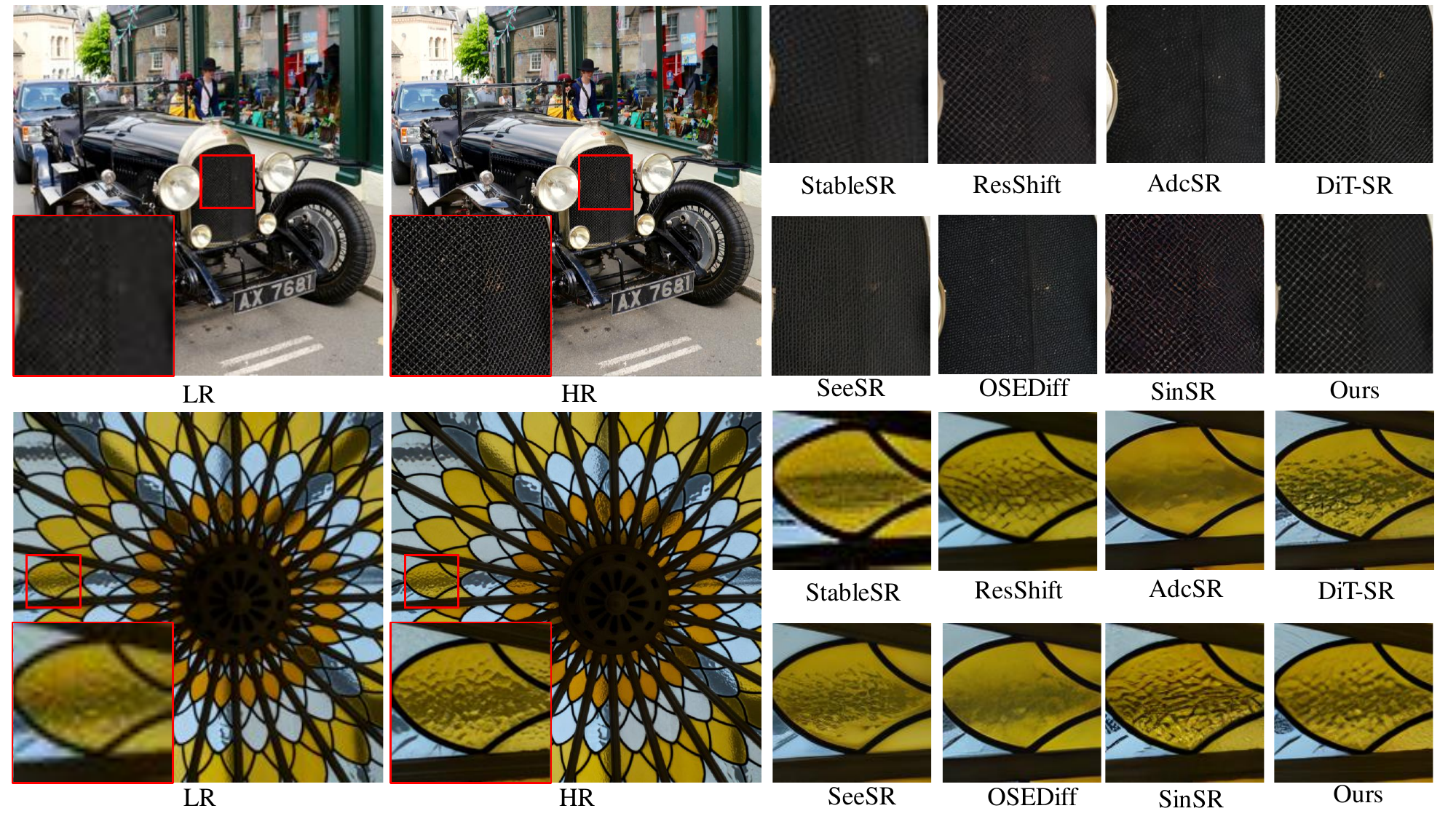} 
\caption{Visual comparisons of different DM-based SR methods on DIV2K.}
\label{fig4-2}
\end{figure*}

The upper part of Table~\ref{tab: quantitative comparison} and Figure~\ref{fig4-2} present the comparison results of HDW-SR with other diffusion-based methods on the DIV2K dataset.ResShift achieves good PSNR and LPIPS results but struggles to recover rich details in generated images. StableSR and SeeSR utilize Stable Diffusion priors and high‑level semantics to recover global structures, but they suffer from noticeable artifacts (e.g., shadows on glass). OSEDiff compresses the diffusion process into one-step inference for greater efficiency, yet its outputs omit many fine details. AdcSR accelerates generation via distillation but produces overly smooth results with weak detail recovery. SinSR performs poorly across most metrics and yields visually unnatural outputs. DiT‑SR benefits from the DiT backbone’s strong capacity, yielding good quantitative scores; however, it tends to over‑enhance textures (e.g., distorted dust grilles on car fronts). In contrast, HDW-SR delivers superior visual quality and structural fidelity. Its lossless wavelet sampling and high‑frequency guidance module align the reconstructed structures closely with the HR images. Additionally, the DTB module further refines the guidance, enabling precise and natural recovery of fine details.

\subsubsection{Comparison Results on Real‑World Datasets}
\begin{figure*}[t]
\centering
\includegraphics[width=0.9\textwidth]{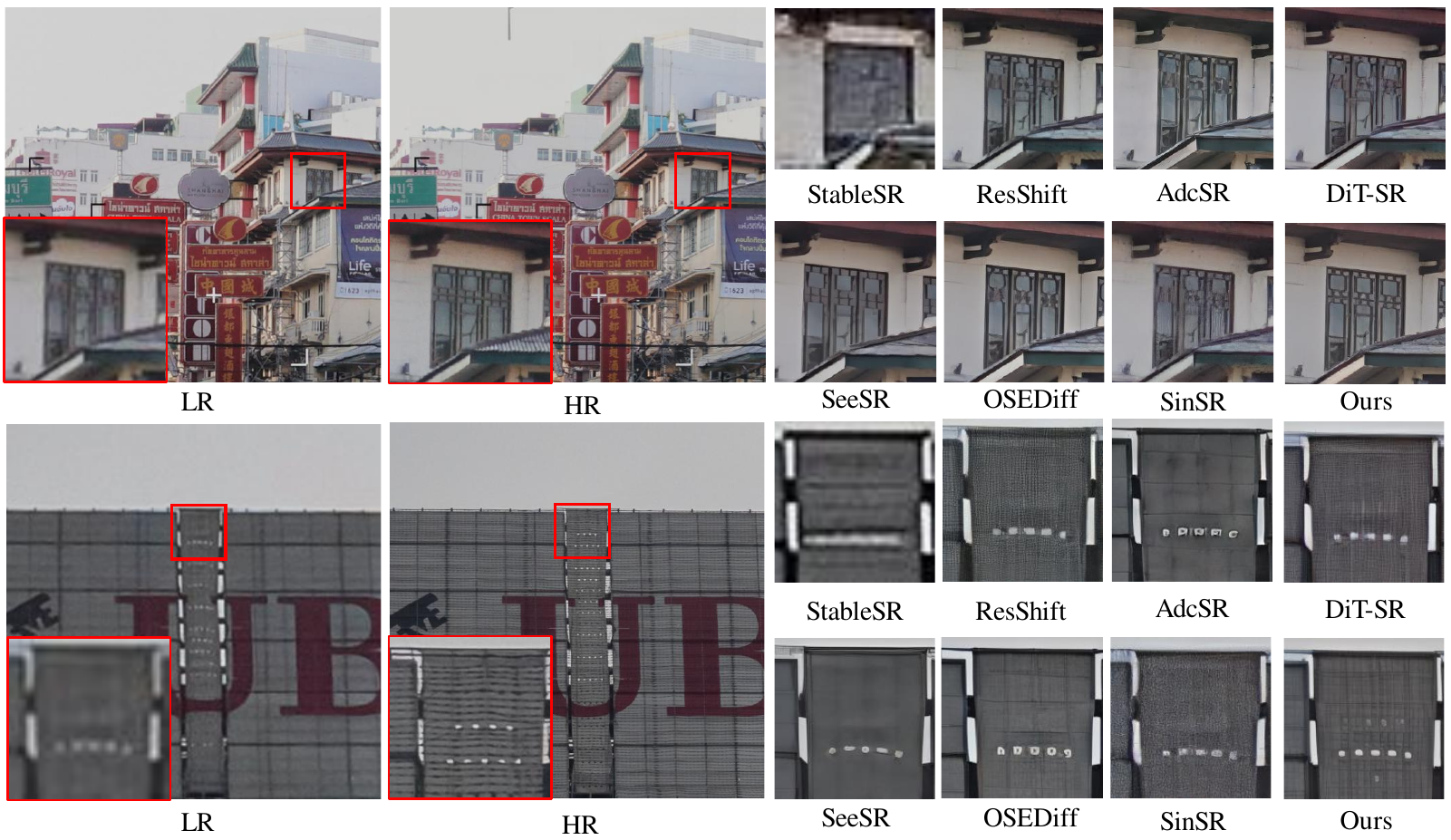} 
\caption{Visual comparisons of different DM-based SR methods on RealSR and DRealSR.}
\label{fig:Visual}
\end{figure*}

The lower half of Table~\ref{tab: quantitative comparison} and Figure~\ref{fig:Visual} compare HDW-SR with other diffusion-based methods on real-world datasets. StableSR and SeeSR exhibit poor performance in both quantitative metrics and visual quality, suffering from noticeable distortions and edge blurring. ResShift still fails to reconstruct rich, coherent details, limiting perceptual quality. Although OSEDiff and AdcSR improve inference speed, they compromise high‑frequency detail restoration (e.g., window contours). DiT-SR and SinSR produce noticeable artifacts in window regions, indicating hallucinations within the network. In contrast, HDW-SR excels in no‑reference metrics and demonstrates superior capability in reconstructing intricate details such as subtle variations (e.g., the cavity in the bottom image of Figure~\ref{fig:Visual}). This advantage is attributed to our high-frequency guidance mechanism and the DTB, which effectively mine and preserve fine-grained image information throughout the diffusion process. 

\subsection{Comparisons with GAN-based Methods}
Table~\ref{tab:comparison} presents a comparison of HDW-SR with three state-of-the-art GAN-based methods: RealESRGAN\cite{wang2021real}, BSRGAN\cite{zhang2021designing}, and LDL\cite{liang2022details}, on both real-world and synthetic datasets. HDW-SR achieves the best performance on no-reference metrics including NIQE, CLIPIQA, and MUSIO, while also demonstrating strong results in PSNR and LPIPS, striking a favorable balance between perceptual quality and content fidelity.

\begin{table}[ht] 
\centering
\scriptsize
\caption{Comparison of HDW-SR with state-of-the-art GAN-based methods on DIV2K, RealSR, and DrealSR datasets.}
\label{tab:comparison}
\resizebox{0.48\textwidth}{!}{ 
\begin{tabular}{ll cccccccc}
\toprule
Datasets & Methods & PSNR$\uparrow$ & LPIPS$\downarrow$ & NIQE$\downarrow$ & CLIPIQA$\uparrow$ &  MUSIO$\uparrow$ \\
\midrule
\multirow{4}{*}{DIV2K} 
    & RealESRGAN    & 24.29 & 0.3112 & 4.68 & 0.5577 & 61.06 \\
    & BSRGAN        & 24.38 & 0.3351 & 4.75 & 0.5091 & 61.20 \\
    & LDL           & 23.83 & 0.3256 & 4.85 & 0.5180 & 60.04 \\
    & HDW-SR        & \textbf{24.52} & \textbf{0.2823} & \textbf{4.43} & \textbf{0.6927} & \textbf{69.68} \\
\midrule
\multirow{4}{*}{RealSR} 
    & RealESRGAN  & 25.69 & 0.2727 & 5.83 & 0.4449 & 60.18 \\
    & BSRGAN      & 26.39 & 0.2673 & 5.66 & 0.5001 & 63.21 \\
    & LDL         & 25.28 & 0.2766 & 6.00 & 0.4477 & 60.82 \\
    & HDW-SR      & \textbf{25.71} & \textbf{0.2672} & \textbf{5.39} & \textbf{0.6702} & \textbf{70.10}\\
\bottomrule
\end{tabular}
}
\end{table}

\begin{table}[t]
\centering
\scriptsize
\caption{Ablation results on the impact of hyperparameter $\beta$ on the DIV2K dataset. Red values indicate the baseline (best) performance. Degradation percentages are approximate and align with observed drops.}
\label{tab:ablation-beta}
\resizebox{0.8\columnwidth}{!}{
\begin{tabular}{l cccc}
\toprule
$\beta$ & PSNR$\uparrow$ & SSIM$\uparrow$ & NIQE$\downarrow$ & CLIPIQA$\uparrow$ \\
\midrule
0.2 & \textbf{24.52}  & \textbf{0.6162}  & \textbf{4.43}  & \textbf{0.6937} \\
0.1 & 18.39  & 0.4622  & 5.76  & 0.4856  \\
0.3 & 23.29  & 0.5854  & 4.65  & 0.6243  \\
0.4 & 22.07  & 0.5546  & 5.09  & 0.5896  \\
0.5 & 10.30  & 0.2835  & 8.86  & 0.2012  \\
\bottomrule
\end{tabular}
}
\end{table}

\subsection{Ablation study}
\subsubsection{Ablation Study on Sensitivity to $\beta$}
To investigate the sensitivity of our model to the hyperparameter $\beta$ (which controls the trade-off between perceptual quality and fidelity in the diffusion process), we conducted an ablation study on the DIV2K dataset. We varied $\beta$ from 0.1 to 0.5 and observed its impact on key metrics. As shown in Table~\ref{tab:ablation-beta}, at $\beta=0.1$, all metrics degrade by 20--30\%; at $\beta=0.3$--$0.4$, the degradation is limited to 5--15\%; and at $\beta \geq 0.5$, the performance collapses, indicating instability in the model. This suggests an optimal range around $\beta=0.2$--$0.3$ for balancing performance.

\subsubsection{Ablation Studies on Components of HDW-Net}

Our ablation studies are conducted under 4× super-resolution on the RealSR dataset to explore the impact of each component in HDW‑Net, by evaluating the effectiveness of the proposed DFA and DTB modules and comparing DWT with CNN. The results of the DWT and DFA validation are shown in Table~\ref{tab:ablation on DFA}. The first and last rows of Table~\ref{tab:ablation on DFA} show that DWT significantly improves network performance compared to standard CNN. The last three rows compare three variants: (1) removing the DFA module and replacing it with a self-attention module, (2) using high-frequency guidance from HA-Net, and (3) generating the guidance internally within HE-Net. The comparison confirms the effectiveness of the high-frequency guidance provided by HE-Net. The results in Table~\ref{tab:ablation DTB} reveal that, compared to the Top-K, DTB better guides high‑frequency information and performs more precise data selection.

\begin{table}[ht]
\centering
\scriptsize
\setlength{\tabcolsep}{10pt}
\caption{Ablation Study of CNN and DFA Models.}
\label{tab:ablation on DFA}
\begin{tabular}{lccc}
\toprule
Methods & PSNR$\uparrow$ & SSIM$\uparrow$ & CLIPIQA$\uparrow$ \\
\midrule
CNN+DFA(HE-Net)    & 25.16 & 0.7252 & 0.6631 \\
DWT+SwinT          & 22.15 & 0.6539 & 0.6127 \\
DWT+DFA(HA-Net)    & 24.39 & 0.6984 & 0.6541 \\
DWT+DFA(HE-Net)    & \textbf{25.71} & \textbf{0.7428} & \textbf{0.6702} \\
\bottomrule
\end{tabular}
\end{table}

\begin{table}[ht]
\centering
\scriptsize
\setlength{\tabcolsep}{8pt}
\caption{Ablation Study of DTB.}
\label{tab:ablation DTB}
\begin{tabular}{lcccc}
\toprule
Methods & PSNR$\uparrow$ & SSIM$\uparrow$ & CLIPIQA$\uparrow$ & FLOPS\\
\midrule
Top-K    & 25.37 & 0.7370 & 0.6629 & 172G \\
DBT    & \textbf{25.71} & \textbf{0.7428} & \textbf{0.6702} & 134G\\
\bottomrule
\end{tabular}
\end{table}

\section{Conclusion}

In this paper, we propose HDW-SR, a diffusion‑based framework for SISR that tackles the challenge of recovering fine details. The HDW‑Net employs wavelet decomposition instead of convolution for downsampling. It also fuses the high‑frequency wavelet components of the PreSR image with the low‑frequency wavelet components of the diffusion noise via sparse cross‑attention, thereby achieving high‑frequency guidance. To further concentrate on fine details, we propose the DTB that dynamically selects the optimal threshold $k^*$ by analyzing the intra- and inter-class variances of elements in the similarity matrix. Moreover, HDW‑Net readily supports wavelet decompositions beyond the three-level configuration, such as four- and five-level decompositions, further enhancing its flexibility. Experimental results demonstrate that the proposed method effectively balances pixel‑level accuracy and perceptual quality, yielding substantial gains in detail reconstruction. 
{
    \small
    \bibliographystyle{ieeenat_fullname}
    \bibliography{main}
}


\end{document}